\documentclass[pdflatex,sn-mathphys]{sn-jnl}

\theoremstyle{thmstyleone}%
\theoremstyle{thmstyletwo}%

\theoremstyle{thmstylethree}%
\usepackage{url}
\begin{document}

\title[Feature Descriptor Matching for Liver Registration]{Learning Feature Descriptors for Pre- and Intra-operative Point Cloud Matching for Laparoscopic Liver Registration}

\author*[1]{\fnm{Zixin} \sur{Yang}}\email{yy8898@rit.edu}

\author[2]{\fnm{Richard} \sur{Simon}}\email{rasbme@rit.edu}

\author[1,2]{\fnm{Cristian} \sur{A. Linte}}\email{calbme@rit.edu }

\affil[1]{\orgdiv{Center for Imaging Science}, \orgname{Rochester Institute of Technology}, \orgaddress{\city{Rochester}, \state{NY}, \country{USA}}} 
\affil[2]{\orgdiv{Biomedical Engineering}, \orgname{Rochester Institute of Technology}, \orgaddress{\city{Rochester}, \state{NY}, \country{USA}}}

\abstract{\textbf{Purpose:} In laparoscopic liver surgery (LLS), pre-operative information can be overlaid onto the intra-operative scene by registering a 3D pre-operative model to the intra-operative partial surface reconstructed from the laparoscopic video. To assist with this task, we explore the use of learning-based feature descriptors, which, to our best knowledge, have not been explored for use in laparoscopic liver registration. Furthermore, a dataset to train and evaluate the use of learning-based descriptors does not exist.

\textbf{Methods:} We present the LiverMatch dataset consisting of 16 pre-operative models and their simulated intra-operative 3D surfaces. We also propose the LiverMatch network designed for this task, which outputs per-point feature descriptors, visibility scores, and matched points.

\textbf{Results:} We compare the proposed LiverMatch network with a network closest to LiverMatch, and a histogram-based 3D descriptor on the testing split of the LiverMatch dataset, which includes two unseen pre-operative models and 1400 intra-operative surfaces. Results suggest that our LiverMatch network can predict more accurate and dense matches than the  other two methods and can be seamlessly integrated with a RANSAC-ICP-based registration algorithm to achieve an accurate initial alignment.

\textbf{Conclusion:} The use of learning-based feature descriptors in LLR is promising, as it can help achieve an accurate initial rigid alignment, which, in turn, serves as an initialization for subsequent non-rigid registration. We will release the dataset and code upon acceptance.
}


\maketitle

\section{Introduction}\label{sec1}

In LLS, pre-operative CT or MRI scans offer precise information about vascular and tumor sites. However, during an intervention, it is challenging for the surgeon to mentally fuse the pre-operative images with the intra-operative laparoscopic images. To mitigate this challenge, image guidance systems \cite{haouchine2013image,collins2020augmented} help surgeons by overlaying the pre-operative information onto the intra-operative scene. One of the crucial components of an image guidance system is the registration, which estimates the transformation between pre- and intra-operative data. In LLR, both 3D-2D \cite{espinel2021using} and 3D-3D registration \cite{modrzejewski2019vivo} methods can be employed; for 3D-3D registration, specifically, the intra-operative 3D surfaces are reconstructed from intra-operative videos \cite{modrzejewski2019vivo} and utilized to constrain the registration solutions.

Registration methods can yield rigid or non-rigid alignment. The rigid registration uses manually or automatically detected landmarks to globally align the 3D pre-operative volume data to the intra-operative 3D surface. To better capture soft tissue deformations, non-rigid registration methods are often needed for final alignment. Non-rigid registration techniques entail two fundamental components: surface matching and volumetric model warping. The former identifies a match between the pre-and the intra-operative surfaces. The latter component uses the surface displacements to deform the volumetric model, so that tumor locations or vascular structures identified in the pre-operative model are correctly mapped to the intra-operative scene \cite{rucker2013mechanics,collins2020augmented}. The volumetric model can also be used as a constraint in the surface matching or after the surface matching estimation. Non-rigid deformations allow many potential solutions; therefore, constraints are needed to limit solutions. Various constraints have been explored to solve the registration problem, including anatomical landmarks  \cite{suwelack2014physics}, contours \cite{collins2020augmented,labrunie2022automatic}, as well as biomechanics-based constraints \cite{plantefeve2016patient,rucker2013mechanics}.

The use of 3D feature descriptors is beneficial because they can provide automatic initialization and constraints for rigid and non-rigid registration.
However, feature descriptors pose several challenges. At first, liver surfaces are very smooth compared to natural scenes, making local features difficult to capture. Second, feature descriptors may not be able to capture global characteristics of the liver because intra-operative data only shows parts of the liver surface. Furthermore, deformations and surface reconstruction noise may negatively affect extracted features as they distort the shapes.

Several handcrafted features \cite{robu2018global,krames2022does} have been studied in liver registration. Although learning-based 3D feature descriptors have been proposed in the computer vision field \cite{li2022lepard,huang2021predator}, they are not designed for LLR and, to our best knowledge, have not been applied in LLR. Most learning-based methods assume the scene is rigid \cite{huang2021predator}; while a few tackle non-rigid cases \cite{li2022lepard}, they assume the principal surface is visible. These two assumptions are often challenged because the liver is globally deformed, and only a small part can be seen in intra-operative data. Pfeiffer \cite{pfeiffer2020non} \textit{et al.} proposed a learning-based biomechanical model to estimate the displacement field of a volume mesh to an intra-operative point cloud. However, this method requires a coarse alignment, often performed manually. Although several public datasets  \cite{suwelack2014physics,modrzejewski2019vivo} have been released, there is still no large public dataset or benchmark available to train and evaluate learning-based methods.

This work explores the use of learning-based 3D feature descriptors for 3D-3D LLR though the following contributions: 1. We describe the generation of a large LiverMatch dataset for studying learning-based 3D feature descriptors in LLR. 2. We propose a learning-based 3D feature descriptor network called LiverMatch for 3D-3D laparoscopic liver registration, which uses a Transformer to obtain self-global and cross-global geometry information from super-points while also predicting per-point feature descriptors from the original point clouds. Furthermore, the network also predicts visibility scores, which help the network focus on the visible pre-operative surface. 3. We evaluate the network relative to another network closest to our proposed network and against a traditional registration method on the dataset.

\section{Methods}\label{sec11}

\subsection{Problem Setting}

We define the point cloud extracted from the surface vertices of a pre-operative liver model as a source point cloud, $\mathbf{S} \in \mathbb{R}^{n\times3}$. The intra-operative point cloud is referred to as the target point cloud $\mathbf{T} \in \mathbb{R}^{m\times3}$, and it is assumed to be generated via a stereoscopic video reconstruction, where $n$ and $m$ are the numbers of points, and $n>m$. Hence, the solution to the pre- to intra-operative registration problem is identifying matches between $\mathbf{S}$ and $\mathbf{T}$.

\subsection{LiverMatch Dataset}
\begin{figure}[thb]
\centering
\includegraphics[width=0.9\textwidth]{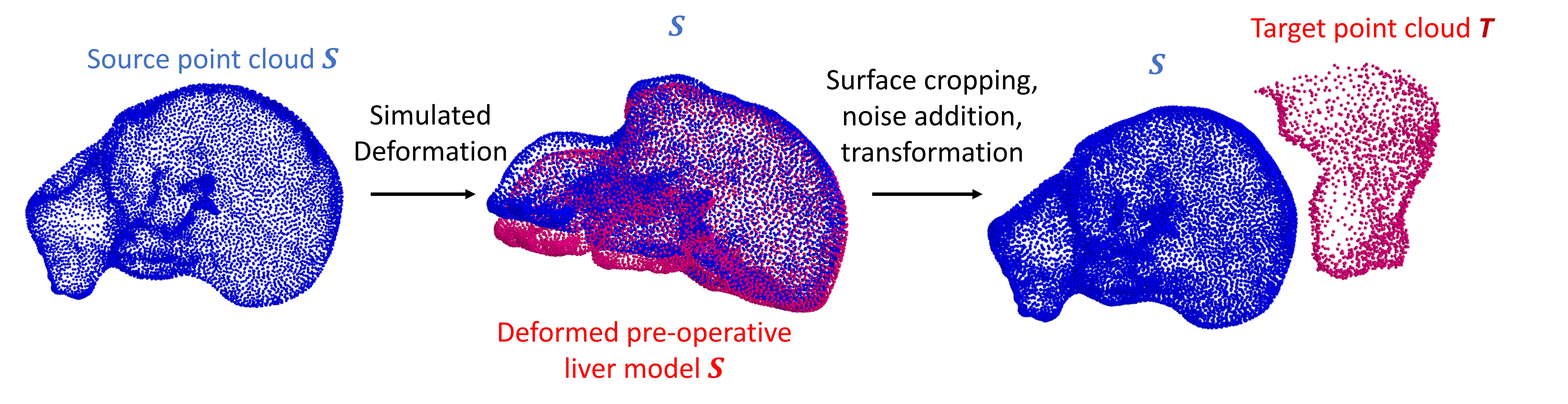}
\caption{Schematic description of the generation of the source ($S$) and target ($T$) point clouds based on 16 liver surface models from the 3D-IRCADb-01 dataset.}
\label{fig:dataset}
\end{figure}

\noindent For this work, the source point clouds are generated from 16 liver models from the 3D-IRCADb-01 dataset\footnote{\url{https://www.ircad.fr/research/data-sets/liver-segmentation-3d-ircadb-01/}} \cite{soler20103d}. The 3D-IRCADb-01 dataset consists of 20 liver models segmented from CT scans. Four liver models (No. 11, 18, 19, and 20) were excluded from this study due to inherent mesh errors. The target, intra-operative point clouds are generated by simulating various deformations of the 3D pre-operative liver surfaces, then extracting different surface regions following deformation. Fig. \ref{fig:dataset} illustrates an example of the generation of the $\mathbf{S}$ and $\mathbf{T}$ point clouds.
 
\textbf{Deformation simulation}. We followed the approach described in \cite{pfeiffer2020non} to generate deformation fields using a neo-Hookean hyperelastic material model with a random Young’s modulus (2 kPa to 5 kPa) and a Poisson’s ratio of 0.35. We applied up to three forces of 3 N maximum magnitude to random surface regions. In addition, random zero-displacement boundary conditions were also prescribed to areas of radius ranging from 15 - 20 mm. These parameters, along with the CT-derived liver geometry and material properties, were input into a finite element solver, which yielded the deformed models. For this study, we selected the deformed regions featuring a 7–15 mm displacement, mimicking deformations similar to those studied using {\it in vitro} phantoms in \cite{suwelack2014physics}.

\textbf{Target point cloud generation}. The following four steps were used to simulate the target point clouds: First, we cropped the front liver surfaces following the simulated deformations and extracted vertices that served as the raw target point cloud. Second, to mimic different visual fields of view of the intra-operative liver surface, we randomly cropped the raw target point clouds using different visibility ratios ($m/n$). Given a raw target point cloud, a 3D unit direction vector was randomly generated; all vertices in the raw 3D point cloud were projected onto the unit vector, and their distances were ranked. We then selected the top points to ensure a visibility ratio between $0.20$ and $0.24$, similar to the visibility ratio of 0.22 achieved in the {\it in vitro} phantom study \cite{suwelack2014physics}. Thirdly, random noise was applied on the cropped point clouds with a maximum magnitude of 2 mm, mimicking accuracy levels similar to those achieved by the state-of-the-art stereo matching methods \cite{edwards2022serv}. Lastly, the point clouds were randomly rotated in all directions and translated by up to 20 mm.

\subsection{LiverMatch Network}

\begin{figure}[thb]
\centering
\includegraphics[width=\textwidth]{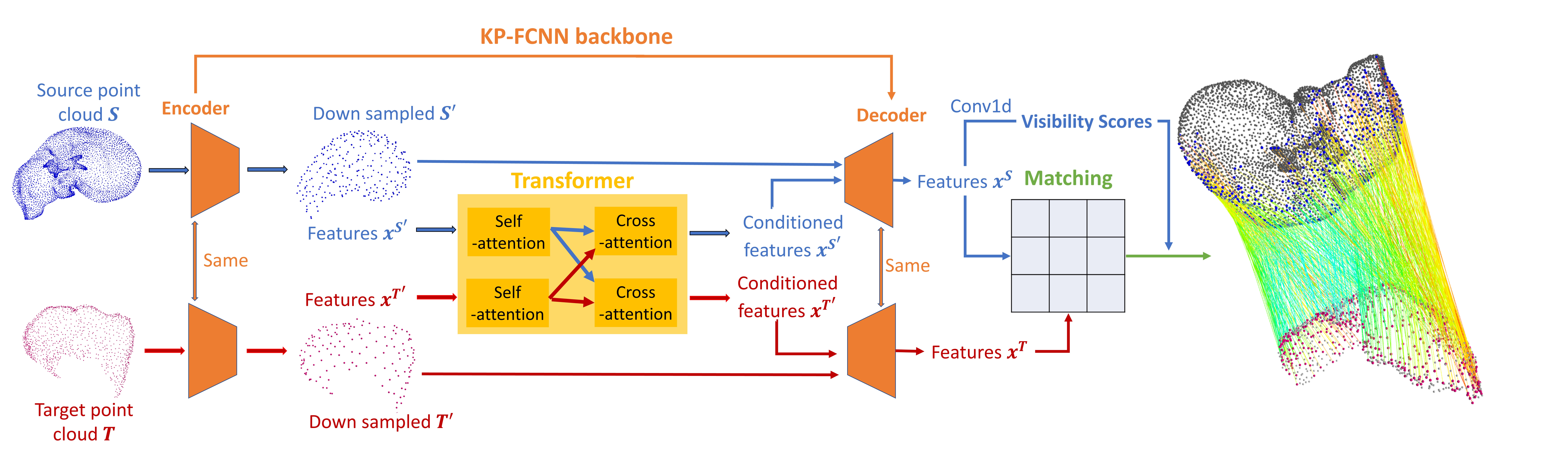}
\caption{LiverMatch network overview: 1) Encoder - down-samples input point clouds and extract associated features; 2) Transformer - updates features to conditioned features with self-global geometry and cross-global geometry information. 3) Decoder - up-samples conditioned features to obtain per-point features. 4) Matching - calculates a confidence matrix to select matches. An additional 1D convolution decodes the $\mathbf{x}^{\mathbf{S}}$ to visibility scores.  }
\label{fig:network}
\end{figure}

\noindent The overview of our LiverMatch network is illustrated in Fig. \ref{fig:network}. The network uses the source and target point clouds as input and outputs point-wise feature descriptors, visibility scores, and matches.

\subsubsection{Encoder}

Given $\mathbf{S}$ and $\mathbf{T}$, the encoder extracts super-points $\mathbf{S'}$ and $\mathbf{T'}$ along with their associated features $\mathbf{x}^{\mathbf{S}'}$ and $\mathbf{x}^{\mathbf{T}'}$. In this network, we use the encoder of the kernel point fully convolutional neural network (KP-FCNN) \cite{thomas2019kpconv}. The encoder consists of ResNet-like blocks and pooling layers based on kernel point convolution (KPConv), which extracts the feature of a point from its neighboring points.

\subsubsection{Transformer}
The features $\mathbf{x}^{\mathbf{S}'}$ and $\mathbf{x}^{\mathbf{T}'}$ only carry information from their close neighborhood points. To overcome the limitation, the single-head Transformer \cite{vaswani2017attention}, consisting of a self-attention layer and a cross-attention layer, is applied to update $\mathbf{x}^{\mathbf{S}'}$ and $\mathbf{x}^{\mathbf{T}'}$ with self-global and cross-global geometry information. The self-attention layer allows points from the same point cloud to communicate, while the cross-attention layer allows points from different point clouds to share information. After the transformer, the features will become conditioned features with self- and global-geometry information. Here, we show an example of updating a source feature $\mathbf{x_i}^{\mathbf{S}'} \in \mathbb{R}^{d \times 1}$ using the self-attention and cross-attention layer:

In the self-attention layer, the query vector $\mathbf{q}$, the key vector $\mathbf{k}$, and the value vector $\mathbf{v}$ are first comupted as:
\begin{equation}\label{eqn:self-attention} 
\mathbf{q}_i =W_\mathbf{q} \mathbf{x}_i^\mathbf{S'}, \;\;\;\;
\mathbf{k}_j = W_\mathbf{k} \mathbf{x}_j^\mathbf{S'},\;\;\;\;
\mathbf{v}_j = W_\mathbf{v} \mathbf{x}_j^\mathbf{S'}, \;\;\;\;
\end{equation}
where $W_\mathbf{q},W_\mathbf{k},W_\mathbf{v} \in \mathbb{R}^{d\times d}$ are learned projection matrices, and $\mathbf{x_j}^{\mathbf{S}'}$ is another source feature. The similarity between $\mathbf{q}$ and $\mathbf{k}$ is measured by:
\begin{equation}
a_{ij}=\text{softmax}(\mathbf{q}_i\mathbf{k}_j^T /\sqrt{d}).
\end{equation}
\noindent $\mathbf{x}_i^{\mathbf{S}'}$ is updated by:

\begin{equation}\label{eqn:transormer_output}
\mathbf{x}_i^\mathbf{S'} =  \mathbf{x}_i^\mathbf{S'} + \text{FC}( \text{Concat}[\mathbf{q}_i,  \sum_{j} a_{ij}\mathbf{v}_j ] ),    
\end{equation}

\noindent where FC(·) denotes a fully connected layer. The same operation is applied to every source feature and target feature.

In the cross-attention layer, $\mathbf{k}$, $\mathbf{v}$ are calculated from the other point cloud. For example, to update the $\mathbf{x_i}^{\mathbf{S}'}$, Eq. \ref{eqn:self-attention} becomes:

\begin{equation}\label{eqn:attention} 
\mathbf{q}_i =W_\mathbf{q} \mathbf{x}_i^\mathbf{S'}, \;\;\;\;
\mathbf{k}_j = W_\mathbf{k} \mathbf{x}_j^\mathbf{T'},\;\;\;\;
\mathbf{v}_j = W_\mathbf{v} \mathbf{x}_j^\mathbf{T'}, \;\;\;\;
\end{equation}

\noindent where $\mathbf{x_j}^{\mathbf{T}'}$ is the feature of the target point cloud. The formations with cross attention are the same after replacing the contents for $\mathbf{q}$, $\mathbf{k}$, and $\mathbf{v}$.

\subsubsection{Feature Decoding}

The conditioned features, along with spatial locations, are then fed to the decoder of KP-FCNN backbone \cite{thomas2019kpconv} to obtain point-wise feature descriptors $\mathbf{x}^{\mathbf{S}}$ and $\mathbf{x}^{\mathbf{T}}$. Following the decoder, we used a 1D-convolution to decode the $\mathbf{x}^{\mathbf{S}}$ into visibility scores $o_{v}$. As only partial points in $\mathbf{S}$ have correspondences, encoding visibility scores helps the network focus on visible points. We clamp the visibility scores within 0 to 1 and create a visibility mask $O_{v} = [ o_{v} > 0.9 ]$.

\subsubsection{Matching}

We first calculate a scoring matrix $S$ and then convert it into a confidence matrix $M$ via the dual-softmax operation \cite{rocco2018neighbourhood,li2022lepard}:

\begin{equation}
S(i,j) = \mathbf{x}^{\mathbf{S}} \cdot (\mathbf{x}^{\mathbf{T}})^{T},   
\end{equation}

\begin{equation}
M(i,j) = \text{Softmax}(\mathcal{S}(i,:))\cdot \text{Softmax}(\mathcal{S}(:,j)),    
\end{equation}

\noindent where $\cdot$ denotes matrix multiplication. Matches are selected from the confidence matrix $M$ via the mutual nearest neighbor criteria: for a pair of matched indexes $(i, j)$, confidence value $\mathcal{M}(i,j)$ should be the maximum value of $\mathcal{S}(i,\cdot)$ and $\mathcal{S}(\cdot,j)$ at the same time. In the end, we use the visibility mask $O_{v}$ to exclude invisible source points.


\subsubsection{Loss Functions}

The total loss $L$ of the network is the sum of two loses: $L=L_M + L_v$, where $L_M$ is the matching loss, and $L_v$ is the visibility loss:

\medskip
\noindent
\textbf{Matching Loss.}
We use the focal loss \cite{lin2017focal} with the default parameters $\alpha=0.25$ and $\gamma=2$ to supervise the confidence matrix $M$:
\begin{equation}
L_M = -\frac{1}{m}\sum_{(i,j)\in K_{gt} } \alpha (1-M(i,j))^{\gamma} \log {M}(i,j), 
\end{equation}

\noindent where $\mathcal{K}_{gt}$ is the set of ground-truth matches with same number $m$ as the target point cloud.

\noindent \textbf{Visibility loss.} We use the binary cross entropy to supervise the visibility scores $o_v$:

\begin{equation}
L_v = -\frac{1}{n} \sum_{i=1}^{n} \hat{o}_{v_i}\log(o_{v_i}) + (1 - \hat{o}_{v_i})\log(1 - o_{v_i}),
\end{equation}

\noindent where $\hat{o}_v$ is the ground truth label, adopting a value of 1 when a source point is visible and 0 otherwise, and $n$ is the number of source points.

\section{Experiments}

We obtained 700 simulated deformations for each liver model. Our proposed network was tested using the No. 1 and No. 2 liver datasets and trained on the remaining 14 datasets. The training data was generated on the fly using pre-simulated deformations and our intra-operative surface generation methods. We generated one target point cloud for each deformation, hence yielding a total of 9800 deformations for training and 1400 for testing. The model was implemented using PyTorch. We used the SGD optimizer with 35 training epochs and a batch size of 1.
Experiments were conducted on a TITAN Xp GPU and an Intel(R) Core(TM) i5-7500 CPU.

\subsection{Evaluation Metrics}

Given a visible source point $\mathbf{S}(i)$, if the predicted correspondence $\mathbf{T}(j)$ is correct, it will lie within a radius $\sigma$ from the ground truth correspondence $\mathbf{T}(\hat{j})$, according to \cite{li2022lepard,huang2021predator}:

\begin{equation}\label{eqn:inlier}
\lVert \mathbf{T}(\hat{j})-\mathbf{T}(j)) \rVert < \sigma.
\end{equation}

\noindent Based on the above definition, an inlier ratio (IR) and a match score (MS) can be calculated to evaluate predicted matches, where higher values indicate a better match.

\textbf{IR} calculates the ratio of the number of inliers $n_{inlier}$ to the number of predicted matches $n_{p}$:

\begin{equation}\label{eqn:ir}
\text{IR} = \frac{n_{inlier}}{n_{p}}.
\end{equation}

\textbf{MS} indicates the ratio of the number of inliers to the number of target cloud points $m$: 

\begin{equation}\label{eqn:ir}
\text{MS} =  \frac{n_{inlier}}{m}.
\end{equation}

If a registration method is employed to estimate displacement vectors for each source point, the registration error (RE) is measured as the root mean square error between the ground truth displacement vectors $\mathbf{V^{gt}}$ and predicted displacement vectors $\mathbf{V^{pred}}$:

\begin{equation}
\text{RE}=\sqrt{\frac{\sum_{i}^{n}\|\mathbf{V}^{gt}(i) - \mathbf{V}^{pred}(i)\|^2}{n}},
\end{equation}
\noindent

\noindent where $n$ is the number of source points, $\mathbf{V^{gt}}$ is the sum of deformation and rigid transformation.

\subsection{Results}

\noindent \textbf{Matching evaluation}. We compared our network with the Predator network \cite{huang2021predator}, the closest method to our proposed framework. To adapt the network to deformable scenes, we used ground truth matches to supervise its loss functions instead of the ground truth rigid transformations. The top-k sampling method in Predator was used to select the best candidate source points to match the target points. Finally, the feature descriptors of selected source points and target points were matched with the same matching method we used in LiverMatch.
\begin{table}[!htpb]
	\centering
	\caption{Evaluation of our LiverMatch against Predator network according to the IR and MS evaluation metrics (mean $\pm$ std dev.) for different inlier radii $\sigma$. *p $<$ 0.05 indicates a statistically significant difference between the LiverMatch and Predator results.}
	\label{tab:eva_cr}
	\resizebox{0.75\linewidth}{!}{
    \begin{tabular}{lcccccc}
			\toprule
            $\sigma (mm)$ & $0$ & $1$ & $2$ & $3$ & $4$ & $5$\\
            \midrule
 			& \multicolumn{6}{c}{\textit{IR\%}} \\
 			\midrule
 			 $^*$Predator\cite{huang2021predator} & 26.82 $\pm$ 4.99 & 26.94 $\pm$ 5.01 &  27.89 $\pm$ 5.10 & 30.61 $\pm$ 5.39 & 35.36 $\pm$ 5.93 & 41.76 $\pm$ 6.60\\
 			LiverMatch & 37.68 $\pm$ 5.91 & 37.91 $\pm$ 5.94 & 39.58 $\pm$ 6.13 & 43.46 $\pm$ 6.61 & 49.21 $\pm$ 7.29 & 55.69 $\pm$ 8.04\\
 			\midrule
 			& \multicolumn{6}{c}{\textit{MS\%}} \\
 			\midrule
 			$^*$Predator$^*$\cite{huang2021predator} & 6.90 $\pm$ 1.84 & 6.93 $\pm$ 1.84 & 7.17 $\pm$ 1.87 & 7.85 $\pm$ 1.95 & 9.04 $\pm$ 2.11 & 10.65 $\pm$ 2.32  \\
 			LiverMatch & 16.95 $\pm$ 3.74 & 17.05 $\pm$ 3.75 & 17.79 $\pm$ 3.83 & 19.50 $\pm$ 4.02 & 22.04 $\pm$ 4.30 & 24.91 $\pm$ 4.64 \\
 			\bottomrule
	\end{tabular}
	}
\end{table}


Table \ref{tab:eva_cr} shows the evaluation of our proposed method against the Predator network in terms of IR and MS for a series of inlier radii ranging from 0 to 5 mm. For a 0 inlier radius, both the IR and MS measure exact matches. Nevertheless, with increasing inlier radius, our method yields higher IR and MS values than the Predator network (p $<$ 0.05), implying our proposed method can predict denser and more accurate matches than the Predator network.  

\noindent \textbf{Ablation study}. We conducted an ablation study on the Transformer and the visibility scores of LiverMatch. When the Transformer was replaced with the graph convolution neural net used in Predator, the IR and MS decreased by 4.96$\%$ and 2.16$\%$, respectively, for $\sigma = 0$. Furthermore, when the visibility scores were removed from the network, both IR and MS dropped by 6.22$\%$ and 3.12$\%$, respectively, also for $\sigma=0$.

\begin{table*}[!t]
\centering
\caption{Assessment of mean feature extraction time (seconds) and RE(mm) upon integration of descriptors with a RANSAC-ICP registration algorithm. *p $<$ 0.05 indicates a statistically significant improvement in the registration achieved using LiverMatch relative to the other descriptor methods.}
\resizebox{0.75\linewidth}{!}{%
\begin{tabular}{|c|c|c|c|}
\hline
& Feature extraction time (s)   & Registration method & RE (mm) \\ \hline
$^*$FPFH\cite{rusu2009fast}&  0.06  & \multirow{4}{*}{RANSAC+ICP}  & 86.28 $\pm$ 49.15  \\ 
$^*$Predator\cite{huang2021predator}& 0.26 & & 8.88  $\pm$   12.19    \\
LiverMatch    & 0.07  &  & 4.83 $\pm$ 3.11  \\
Ground Truth    & -  &  & 3.89 $\pm$ 2.08  \\
\hline
\end{tabular}
}
\label{tab:registration}
\end{table*}
\begin{figure}[h]
\centering
\includegraphics[width=0.85\textwidth]{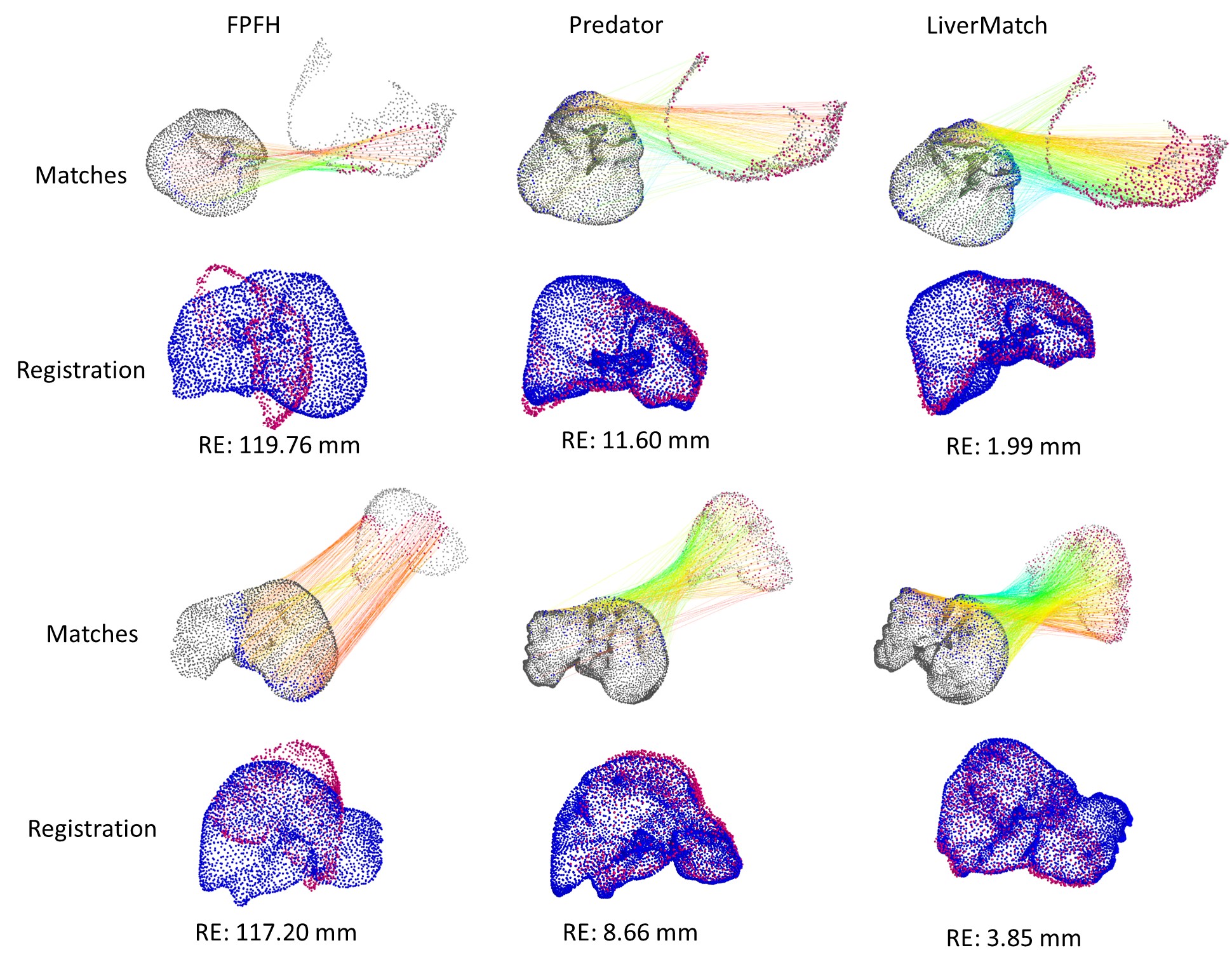}
\caption{Visualization of matches and registration results of FPFH, Predator, and LiverMatch on two pairs of the source point cloud (blue) and target point cloud (red). The first two rows show the results for the same pair of source and target point clouds, while the last two show the results for another pair of source and target point clouds. Unmatched points are shown in gray (the first and third rows). Note that point clouds in FPFH are down-sampled.}
\label{fig:compare}
\end{figure}

\noindent \textbf{Registration evaluation}. We investigated the integration of learning-based point cloud matching with a RANSAC ICP (iterative closest point) rigid registration algorithm. Table \ref{tab:registration} summarizes the registration results achieved using the Fast Point Feature Histograms (FPFH) descriptors \cite{rusu2009fast}, the learning-based matching point cloud descriptors (Predator and our LiverMatch ), and ground truth matches. As FPFH requires heavily down-sampled point clouds, we used a voxel size of 5 mm to downsample the source and target point clouds. However, ground truth correspondences were lost after voxelization, so we could not report the IR and MS scores for FPFH.  We used the Open3D implementations\footnote{\url{http://www.open3d.org/docs/release/tutorial/pipelines/global_registration.html}} of RANSAC ICP and FPFH. 

As shown in Table 2, the integration of RANSAC-ICP with learning-based matching point cloud descriptors (via Predator and LiverMatch) outperforms the FPFH approach. Moreover, our proposed LiverMatch framework yields the lowest registration error (4.83 $\pm$ 3.11 mm), which is comparable to the ground truth registration error of 3.89 $\pm$ 2.08 mm and indicates a statistically significant (p $<$ 0.05) registration improvement over both Predator and FPFH. The achieved registration results are based on a rigid ICP registration and suggest that a non-rigid registration is needed to further reduce registration error. Lastly, LiverMatch yielded a mean feature extraction time of 0.07 s which was comparable to the mean feature extraction time of FPFH (0.06 s) and much shorter than that of the Predator network (0.27 s).

Fig. \ref{fig:compare} illustrates two cases of the point cloud matching and registration results. The target point cloud is occluded in the first case (first two rows). For this challenging case, the learning-based methods can still predict accurate and dense matches. However, FPFH does not yield correct matches, resulting in high registration errors for both cases.

\section{Discussion}\label{sec13}

To generate the LiverMatch dataset, we set limits on the deformation displacements, visibility ratios, and noise magnitude. However, the robustness of learning-based methods to the above factors is still the subject of our ongoing research. Nevertheless, our experiments suggest that our LiverMatch network can find accurate and dense matches between pre- and intra-operative point clouds. Furthermore, the predicted matches can be integrated into rigid-registration methods to achieve fast and accurate rigid alignment.

We tested the Predator and our LiverMatch network on "unseen" liver datasets and their simulated target point clouds. However, the performance of these methods in the clinical setting is still unclear, as it has not been assessed. In addition, whether the learning-based feature descriptors trained on arbitrary objects can generalize to different organs, as speculated in \cite{pfeiffer2020non}, has yet to be further investigated.

On the other hand, Li and Harada \cite{li2022lepard} proposed a network that includes a Transformer that features a repositioning technique; however, when implementing their approach, the training loss did not converge when the network was trained on our LiverMatch dataset, and, furthermore, the network cannot predict per-point features on original, native resolution point clouds.

We also demonstrated the integration of point cloud matching from learning-based feature descriptors with a rigid ICP registration algorithm. It demonstrated rapid feature extraction and comparable results to ground truth registration. We will further investigate the integration of point cloud matching with a non-rigid registration method. Specifically, we will research a non-rigid registration method that can handle dense matches, outliers, and noisy target surfaces, to more closely mimic typical clinical datasets.

Nevertheless, while acknowledging several limitations and ongoing research efforts discussed here, to our knowledge, this paper constitutes the first investigation of using learning-based feature descriptors for laparoscopic liver registration showing, and it features several promising results, including compelling matching performances, extraction times, and registration results upon integration with rigid ICP.

\section{Conclusion}

In this paper, we have presented the generation of the LiverMatch dataset to enable us to study the use of learning-based matching descriptors for laparoscopic liver registration, as well as introduced the LiverMatch network that was shown to yield accurate and dense pre- to intra-operative surface matches. Our results suggest that the use of learning-based descriptor matching in conjunction with laparoscopic liver registration is promising, as it not only offers a rapid and accurate rigid alignment of the pre-and intra-operative liver surfaces but also has the potential to assist with non-rigid registration.

\bibliography{sn-bibliography}

\end{document}